\documentclass[final,times,sort&compress]{elsarticle}

\usepackage{lineno,hyperref}
\modulolinenumbers[5]

\usepackage{longtable}
\usepackage{tabulary}
\usepackage{graphicx}

\usepackage{subfig}
\usepackage [english]{babel}
\usepackage [autostyle, english = american]{csquotes}
\MakeOuterQuote{"}

\usepackage{listings}
\usepackage{enumerate} 
\usepackage[section]{placeins}
\usepackage{amsmath}

\usepackage{booktabs}
\usepackage{multirow}
\usepackage{float}

\usepackage{fancyhdr}
\usepackage{textcomp}

\usepackage{threeparttable}

\fancypagestyle{plain}{%
  \fancyhf{}
  \fancyfoot[C]{\iffloatpage{}{\thepage}}
  }
\journal{ a Journal}
 
\begin{document}

\begin{frontmatter}	

\title{A BERT based Ensemble Approach for Sentiment Classification of Customer Reviews and its Application to Nudge Marketing in e-Commerce}


%

\author[myu]{Sayan Putatunda \corref{cor1}}

\address[myu]{Walmart Global Tech}

\cortext[cor1]{Corresponding author}

\author[myu]{Anwesha Bhowmik}

\author[myu]{Girish Thiruvenkadam}

\author[myu]{Rahul Ghosh}


\begin{abstract}
According to the literature, Product reviews are an important source of information for customers to support their buying decision. Product reviews improve customer trust and loyalty. Reviews help customers in understanding what other customers think about a particular product and helps in driving purchase decisions. Therefore, for an e-commerce platform it’s important to understand the sentiments in customer reviews to understand their products and services, and it also allows them to potentially create positive consumer interaction as well as long lasting relationships. Reviews also provide innovative ways to market the products for an e-commerce company. One such approach is “Nudge Marketing”. Nudge marketing is a subtle way for an e-commerce company to help their customers make better decisions without hesitation. 

The objective of this paper is to build a sentiment analysis model based on customer reviews for a few particular product types of an e-commerce platform. This will help the organization to determine and categorize opinions about a product. We use deep learning and natural language processing techniques to process the reviews text data and develop models for sentiment (positive, negative, or neutral) classification using techniques such as bi-directional LSTMs (Long Short Term Memory) and BERT (Bidirectional Encoder Representations from Transformers). We then proposed two different Stacking/ensembling approaches. We perform our experiments on multiple real-world customer reviews data from different e-commerce platforms. We find that one of the proposed stacked ensemble approach for sentiment classification has a better prediction accuracy. Finally, we provide some illustrations of how the output from this model can be applied for Nudge marketing by e-commerce firms to help their customers in making purchase decisions. We also perform a k-armed Bandit Experiment for the Review based Nudging Strategy using Thompson sampling and Monte Carlo simulation. From the bandit experiments we conclude that the review based nudge/badge seems to be an effective strategy either as a standalone nudge or as a combination with other nudges.

\end{abstract}

\begin{keyword}
Transformers \sep BERT \sep Deep Learning \sep NLP \sep Reinforcement Learning \sep Nudge Marketing \sep Sentiment Analysis
\end{keyword}

\end{frontmatter}


\section{Introduction} \label{intro}
More consumers are reading online reviews than ever before. According to some recent marketing research reports \citep{rv:1}, around $91\%$ of people read them and $84\%$ trust them as much as they would a personal recommendation. In fact, customers like to see multiple reviews and give almost equal weightage to both positive and negative reviews in their decision making. Customer reviews have changed and created innovative ways for e-commerce companies to market to them. Reviews allow e-commerce firms to understand what their customers think of their products, service as well as themselves. It allows them to potentially create positive consumer interaction as well as long lasting relationships. 

In sum, some of the benefits of positive customer reviews for an e-commerce firm are increased sales, build trust, contribute to SEO efforts, and aid customer decision-making. Reviews help customers in understanding what other customers think about a particular product and helps in driving purchase decisions. Therefore, for an e-commerce platform it’s important to understand the sentiments in customer reviews to understand their products and services, and it also allows them to potentially create positive consumer interaction as well as long lasting relationships. Sentiment analysis is contextual mining of text which identifies and extracts subjective information in source material, and helping an e-commerce business to understand the social sentiment of their brand, product or service while monitoring online customer reviews. In other words, Sentiment Analysis is a text classification tool (i.e. a Natural Language Processing technique) that analyses an incoming message (Reviews in the context of this paper) and tells whether the underlying sentiment is positive, negative, or neutral. 

The contributions of this paper is in the development of a BERT based embedded ensemble approach for sentiment classification of customer reviews in an e-commerce platform. We use deep learning and natural language processing techniques to process the reviews text data and develop models for sentiment (positive, negative or neutral) classification using techniques such as LSTMs (Long Short Term Memory), Glove, and BERT (Bidirectional Encoder Representations from Transformers). We perform our experiments on three real-world customer reviews data from different e-commerce platforms. One dataset is an internal data source from Walmart.com for a particular product type- "Headphones" and the other two datasets are publicly available. We compare our proposed solution i.e. BERT based embedded ensemble approach with the benchmark standalone techniques mentioned earlier.  We find that the proposed approach for sentiment classification has a better prediction accuracy. Finally, we provide some illustrations of how the output from this model can be applied for Nudge marketing by e-commerce firms to help their customers in making purchase decisions. We also perform two k-armed Bandit experiments for the Review based Nudging Strategy and combination of nudges strategy using Thompson sampling and Monte Carlo simulation. From the bandit experiments we conclude that the review based nudge/badge seems to be an effective strategy either as a standalone nudge or as a combination with other nudges.

The rest of this paper is structured as follows. In Section \ref{lit} we give a brief overview of the literature on Sentiment Classification of customer reviews using machine learning and deep learning.  This is followed by a discussion on the relevant problem (along with data description)  in Section \ref{prob}. Section \ref{eval} describes the proposed solution and the evaluation metrics. In Section \ref{result1}, we discuss the experimental results and Section \ref{nudge} describes the applications of the proposed solution to Nudge marketing. Then in Section \ref{nudge1}, we discuss two stochastic k-armed bandit tests with Thompson sampling and Monte Carlo simulation for different review based nudging strategies. Finally, Section \ref{conl} concludes the paper.

\section{Related Work} \label{lit}
The literature is replete with studies that proposed Sentiment Analysis based on machine learning \citep{sayanbook:1} and deep learning \citep{sayanbook:2}.  Manek et al. \citep{rv:3} used support vector
machine (SVM) for classification and performed feature extraction based on a gini index. Singh et al. \citep{rv:5} used four different machine learning algorithms for text sentiment analysis such as, J48, naive bayes, BFTree, and OneR. Hai et al. \citep{rv:4} proposed a new probabilistic
supervised method that could infer the overall sentiment of the comment data.

Hyun et al. \citep{rv:7} proposed a
target-dependent convolutional neural network. Jianqiang et al. \citep{rv:6} used the contextual semantic features and
the co-occurrence statistical features of the words in the tweet
and the n-gram feature input convolutional neural network to
analyze the sentiment polarity.  There have been quite a few works on LSTMs (Long Short-Term Memory Networks) reported in the literature. For example, Chen et al. \citep{rv:8} proposed
the LSTM model for the detailed emotional analysis of Chinese product reviews. Hu et al. \citep{rv:9} performed sentiment analysis of short texts by
constructing a keyword vocabulary and combining the LSTM
model. Durairaj and Chinnalagu \citep{rv:10} used Bidirectional Encoder Representations from
Transformers (BERT) to build a contextual model for sentiment analysis of customer reviews.

\section{Problem Statement \& Data} \label{prob}
In this paper, we are looking at a supervised Sentiment Classification problem. This is a multi-class classification problem where the independent variables are text features such as Review and the target variable is 'Sentiment' that contains three classes namely, Positive, Negative, and Neutral. 

In this paper, we use a Walmart's dataset where we take the review data for a particular category i.e. Headphones. The total number of observations are around 86k. The dataset contains review headline, review text and the target labels. The dataset is labelled using user rating. From our previous experimentations we have seen that user rating and review sentiment are positively correlated. So, we classify the review sentiments into three categories, POSITIVE (ratings 4 and 5), NEGATIVE (ratings 1 and 2) and NEUTRAL (rating 3). We will refer to this dataset as "Data 1" for the rest of this paper.


One of the other datasets that we have taken contains app reviews for a fashion item oriented e-commerce firm. The dataset is publicly available at Kaggle \citep{nykaa}. This is a labelled dataset containing reviews text as the independent variable and the target variable contains three classes namely, Positive, Negative, and Neutral. This dataset contains around 200k observations. We will refer to this dataset as "Data 2" for the rest of this paper.

The final dataset used in this paper is the product review data for the category- "Clothing, Shoes and Jewellery" for a global e-commerce firm. This dataset contains product reviews and metadata, spanning May 1996 - July 2014 \citep{amazon:1}. This dataset contains around 278,677 observations. We will refer to this dataset as "Data 3" for the rest of this paper.

\section{Evaluation Metrics} \label{eval}

In this paper, we are going to use the F1-Score and the Prediction Accuracy on the validation dataset as evaluation metrics. The F-score is a way of combining the precision and recall of the model, and it is defined as the harmonic mean of the model's Precision and Recall \citep{fscore}. The higher the F1-Score, the better the performance of the classification method. Similarly, higher the Prediction Accuracy, the better the performance of the classification method. We will also record other metrics such as, Precision and Recall.

\section{Proposed Solution and Experimental Results} \label{result1}
 
In this paper, we first use different kinds of word embeddings for different models. Some of the word embeddings techniques used in this paper are (a) BERT (Bidirectional Encoder Representations from Transformers) based word embedding \citep{bert}, (b)	Glove based word embedding followed by document pool embedding \citep{glove}, and (c) LSTM embeddings \citep{lstm}. We then use a Text classifier for model training and prediction. The text classifier takes different word embeddings and then classify the text representations into labels. As mentioned earlier in Section \ref{prob}, we are solving  a multi-class classification problem i.e., sentiment analysis on reviews data. We have used one Walmart dataset and another external dataset (see Section 3). The datasets contain review text as independent variable and the target variable contains three classes namely, Positive, Negative, and Neutral. Once we get the data after pre-processing, we check how many samples are present for different categories. To address the issue of class imbalance, we perform a down sampling to extract a balanced dataset from the raw data. Followed by this, we perform a stratified sampling of the data to split it into train and test data where, $80\%$ is train and $20\%$ is test data. We then apply the three different embeddings as mentioned above and perform text classification after performing hyper-parameter tuning. We will refer to these three standalone models as BERT, Glove, and LSTM for the rest of this paper. All the experiments were performed in a system with configuration of 16 GB RAM, Apple M1 chip, and Mac OSX. 


We have developed an ensemble model to predict the classes in the test datasets by stacking the three standlone models- BERT, LSTM, and Glove. For this we are using two methods:\\
(1)	We are using majority voting method for predicting labels. If two of the above-mentioned model outputs agree on some labels that label is chosen for test set data prediction. If three models are giving completely different labels, we will be using BERT based labelling.  We will refer to this method as Stack1 for the rest of this paper. \\
(2)	Here we are creating a layer at the end of initial model computation where we take the outputs from the three models and make them as features for the logistic regression. The initial three models are the base models and logistic regression is acting as a meta learner. Finally, the logistic regression model is being used for prediction in the test set. We will refer to this method as Stack2 for the rest of this paper.

In Table \ref{tab1}, we describe the experiment results of all the methods discussed above on the three datasets. In the first dataset i.e. Data 1, we find that among the standalone methods, BERT is the best performer with a prediction accurracy of $0.82$ and F1-score of $0.73$. But one of the proposed ensemble approach i.e. Stack2 is the best performer with with a prediction accurracy of $0.87$ and F1-score of $0.83$. For Data 2, both BERT and Stack2 are the best performing methods with a prediction accurracy of $0.70$ and F1-score of $0.69$. And finally in the third dataset i.e. Data 3, we find that both the ensemble models i.e. Stack1 and Stack2 perform slightly better than the BERT model.

\begin{table*}[!htp]
\caption{Results of various machine learning models on the three different datasets (The test data used is the random 20\% split of the original dataset)}
\label{tab1}
 \scalebox{0.99}{
\begin{tabular}{cccccc}
\hline
Dataset                  & Methods & Accuracy & Precision & Recall & F1-Score \\ \hline
\multirow{5}{*}{Data 1} & BERT    & 0.82     & 0.75      & 0.72   & 0.73     \\
                         & Glove   & 0.55     & 0.38      & 0.32   & 0.35     \\
                         & LSTM    & 0.79     & 0.61      & 0.53   & 0.56     \\
                         & Stack1  & 0.83     & 0.59      & 0.55   & 0.56     \\
                         & Stack2  & 0.87     & 0.84      & 0.82   & 0.83     \\ \hline
\multirow{5}{*}{Data 2}   & BERT    & 0.70      & 0.69      & 0.70    & 0.69     \\
                         & Glove   & 0.38     & 0.46      & 0.28   & 0.33     \\
                         & LSTM    & 0.65     & 0.53      & 0.48   & 0.50      \\
                         & Stack1  & 0.66     & 0.52      & 0.50    & 0.50      \\
                         & Stack2  & 0.70      & 0.69      & 0.70    & 0.69    \\ \hline
\multirow{5}{*}{Data 3}   & BERT    & 0.69      & 0.70      & 0.69    & 0.69     \\
                         & Glove   & 0.31     & 0.50      & 0.23   & 0.31     \\
                         & LSTM    & 0.58     & 0.52      & 0.43   & 0.58      \\
                         & Stack1  & 0.70     & 0.70      & 0.69    & 0.70      \\
                         & Stack2  & 0.70      & 0.70      & 0.69    & 0.70    \\ \hline
\end{tabular}}
\end{table*}

\section{Application to Nudge Marketing} \label{nudge}
Reviews also provide innovative ways to market the products for an e-commerce company. One such approach is “Nudge Marketing”. Nudge marketing is a subtle way for an e-commerce company to help their customers make better decisions without hesitation. Nudge marketing can be defined as the process of communicating marketing messages that encourage desired behaviour by appealing to the psychology of the individual. This is derived from Richard Thaler and Case Sunstein’s Nudging principle, where the authors define Nudge as- “ any aspect of the choice architecture that alters people’s behaviour in a predictable way without forbidding any options or significantly changing their economic incentives.” \citep{nudge:1}

\begin{figure*}[!htp]
\centering
\includegraphics[width=0.9\textwidth]{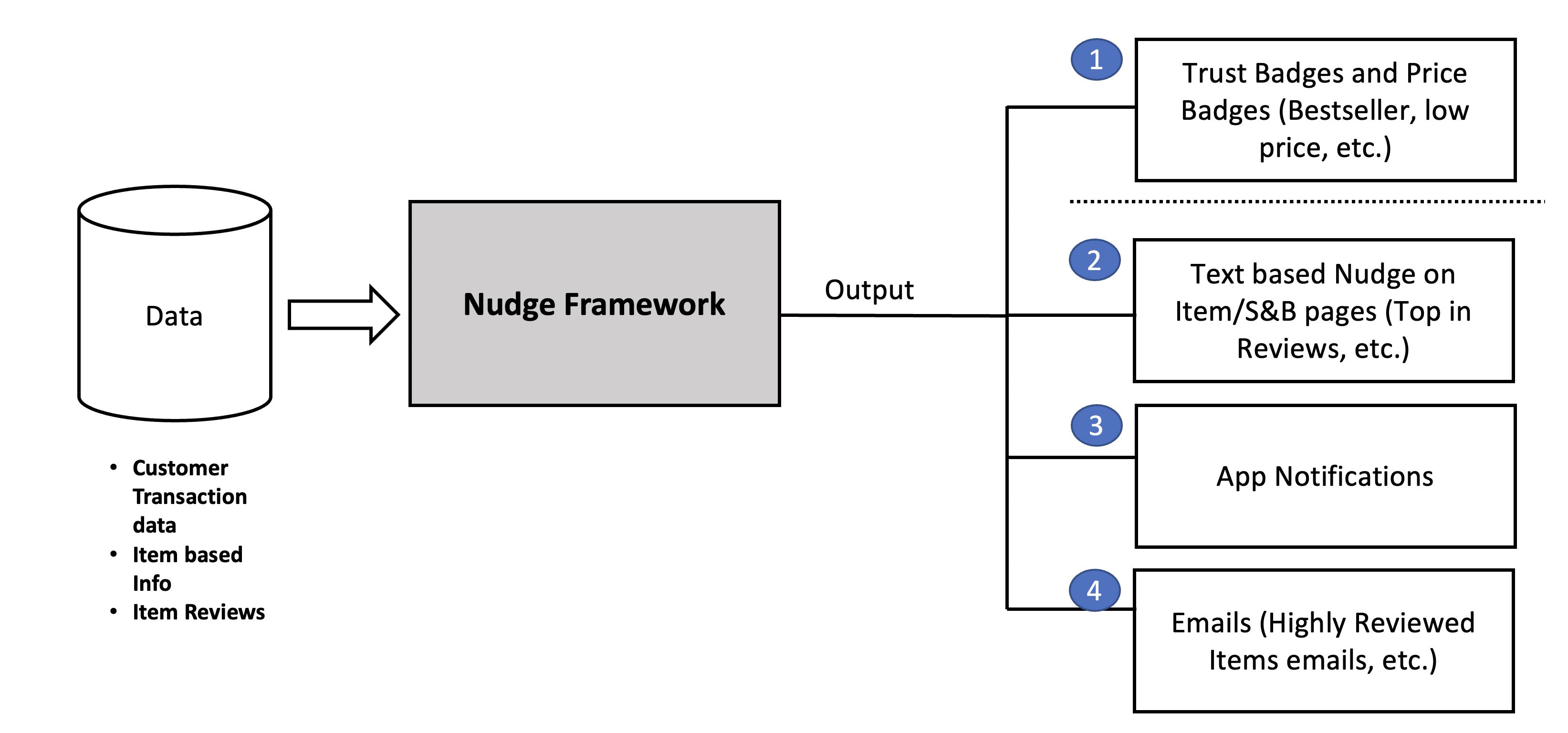}
\caption{Nudge Framework- Flow Diagram}
\label{fig:node1}
\end{figure*}

The idea of nudging is to help customers in making their purchase decisions and in the process helping the e-commerce firms in attaining its business objectives. There are various ways an e-commerce organization can deliver these behavioural nudges to their customers. In Figure \ref{fig:node1}, we show the nudge framework flow diagram. The input to this framework is information from various data sources such as, customer transaction data, item based features data, reviews data, and more. Then we have an abstract layer that contains various machine learning models and rule-based engines to process the input data and give the output. The Output part shown in Figure \ref{fig:node1}, are basically the various forms of nudges. The first one is Trust badges or Price badges. Some examples of trust badges that can be seen as tags on an item image across various e-commerce websites are "Bestseller", "Most Popular", "Guiltfree", etc.


However, in the context of this paper, we will focus on the other types of nudges shown in Figure \ref{fig:node1} (nudges 2, 3, and 4).

\begin{figure*}[!htp]
\centering
\includegraphics[width=0.9\textwidth]{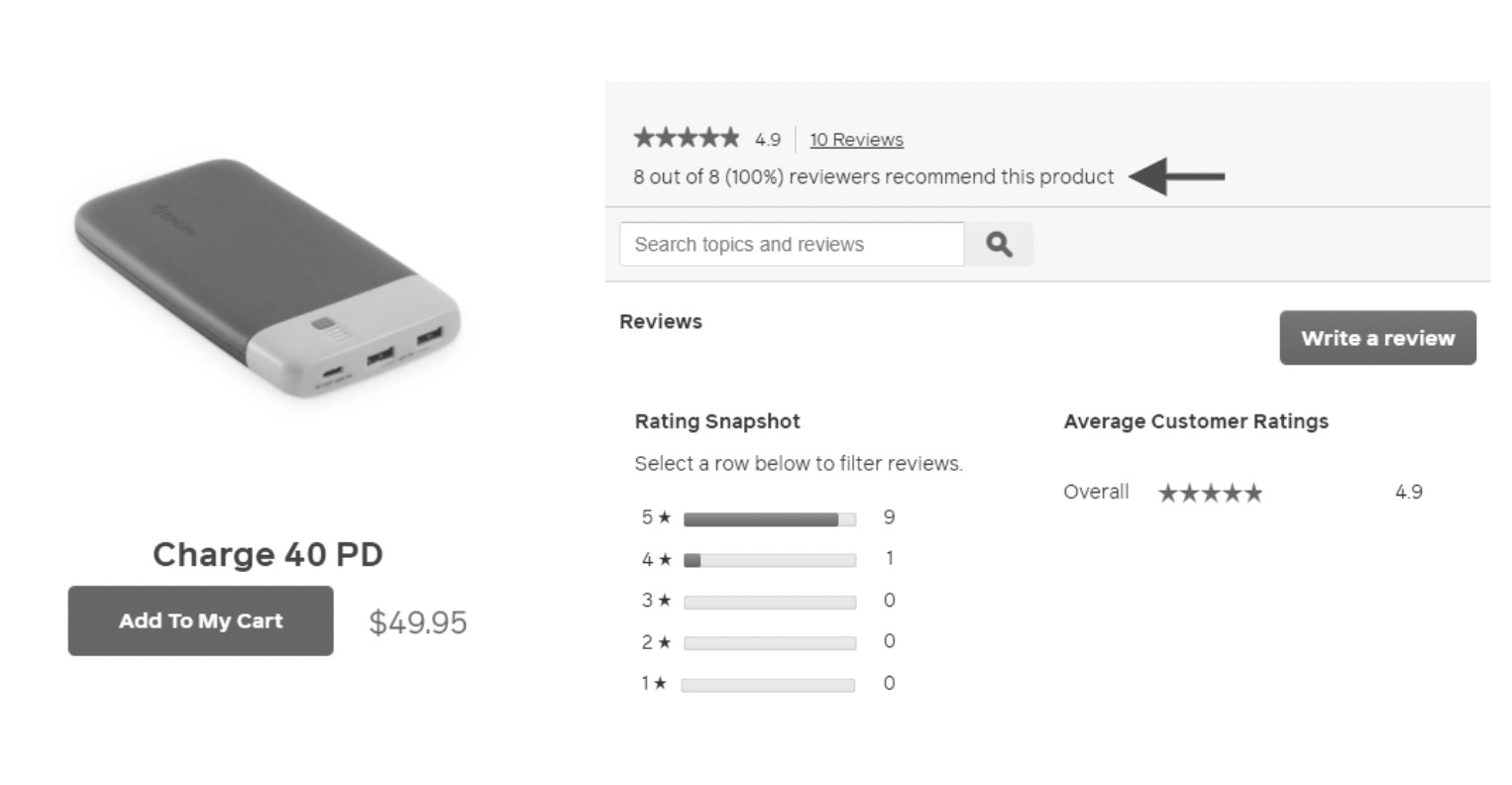}
\caption{An example of a Review based Nudge (Image Source: Biolite \citep{random:1})}
\label{fig:node3}
\end{figure*}

The other types of nudges shown in the output part of Figure \ref{fig:node1} are Text based Nudges, app notifications, and targeted emails. For the purpose of application of nudge marketing, the Data 1 and the Data 3 datasets discussed in Section \ref{prob}, are more relevant in this context as they contain reviews data of items for a product type (such as, Headphones or clothing). We apply the proposed ensemble approach i.e. stack2 discussed in Section \ref{result1} for Sentiment Classification of customer reviews of various items. We then use the model output to identify a set of items that have a high positive sentiment based on reviews and we can then use it to give targeted nudges to the customers. Figure \ref{fig:node3} shows an example of a review based nudge taken from the Biolite's website, where the text "8 out of 8 ($100\%$) of reviewers recommend this product" appears at the top of the item page. In the context of our use-case, we can use text based nudges such as, "x out of y people have highly rated this", "Top in Reviews", and more. We can also provide nudges to a targeted set of customers (based on their transaction history) via app notifications and emails for highly rated/reviewed items. The sentiment classification model needs to be updated regularly and the performance of the nudges need to be monitored using various metrics such as, CTR (Click through rate), lift in Sales, Weekly Active Users (WAU), and more.

In the next section (i.e., Section \ref{nudge1}), we discuss two stochastic k-armed bandit tests with Thompson sampling and Monte Carlo simulation for the review based nudging strategy and for the combination of nudges strategy.  

\section{Multi-armed Bandit Experiment for Nudging Strategy using Thompson sampling and Monte Carlo simulation} \label{nudge1}
So far, we have discussed how we can use the model output to identify a set of items that have a high positive sentiment based on reviews and then use it to give targeted nudges to the customers by showing badges (as shown in Figure 3). One of the standard way in the e-commerce industry to test a new feature (in this case a review based badge/nudge) is to perform an AB test.  The test involves two groups- a treatment group (who have access to the new feature) and a control group. Then we measure a key metric for the two groups such as, click through rate (CTR), and more. The difference between the groups is then tested for statistical significance. However, conducting an AB test is an expensive and a time consuming process. Another disadvantage of AB test is that if the treatment group is clearly superior, we still have to spend lots of traffic on the control group, in order to obtain statistical significance.  Taking these factors into account, in the context of this paper, we perform a multi-armed bandit experiment for nudging strategy in this section. This is something similar to what Geng et. al. \citep{thom:3} and Mao et. al. \citep{thom:2} had done in different contexts i.e., they used a multi-armed bandit approach for online advertising testing and for news headline testing respectively. However, the literature doesn't contain many studies (to the best of our knowledge) where a multi-armed bandit approach has been used to test the effectiveness of badges/nudges in an e-commerce platform.

In the context of this paper, we perform a multi-armed bandit experiment, the foundation of which is Bayesian updating. Here, each arm has a probability of success, which is modeled as a Bernoulli process. The probability of success is unknown, and is modeled by a Beta distribution. As the experiment continues, each arm receives user traffic, and the Beta distribution is updated accordingly.

Here, we will perform our experiments on a particular category for an e-commerce firm. In the first experiment, we have k=2 arms. Each arm is a nudging strategy with click-through rate (CTR) that follows a Beta distribution. We calculate the CTR based on the historical interaction data. Please note that due to confidentiality reasons, we cannot quote the exact CTR numbers. So, all the CTR numbers quoted below are approximate figures. The purpose here is to run a simulation and show the effectiveness of different nudging strategies. The arms are defined as follows:\\
1) Arm 1- This represents the current experience in this particular e-commerce firm's website for the concerned category. By current experience we mean that there is no review based badges in the website. But there are some trust badges (such as Bestseller, etc.) and price based badges as shown in figure 1. The overall  CTR (click through rate is defined as sum of number of clicks divided by the sum of number of impressions) i.e. taking into account both badges and non-badges items in the concerned category is say, $0.021$. We calculate the CTR by checking the clicks and impressions data for all the items in the category for a specific time period.  \\
2) Arm 2- This represents the review based nudging strategy. Here, we identify set of items eligible for the "Top in Reviews" badge by using the sentiment classification model (Stack2) discussed in Section 5 i.e. the items with Positive sentiment labels are taken into account. We then check the CTR for these set of items from the data and let's assume, it is around $0.024$. Here, the assumption is that the CTR is going to stay the same even if a "Top in Reviews" badge badge is visible to customers for these set of items.    \\
The goal of the experiment is to find the nudging strategy with the highest click through rate.

We use Thompsom sampling, which is a greedy method that always chooses the arm that maximizes expected reward. In each iteration of the bandit experiment, Thompson sampling simply draws a sample CTR from each arm’s Beta distribution, and assign the user to the arm with the highest CTR. Here, Thompson sampling and Bayesian update work hand-in-hand. If one of the arms is performing well, its Beta distribution parameters are updated to remember this, and Thompson sampling will more likely draw a high CTR from this arm. Throughout the experiment, high-performing arms are rewarded with more traffic, whereas under-performing arms are punished with less traffic.

However,  the Beta distribution estimates the CTR, we need to know how confident we are about each estimate of CTR. If we are confident enough about the arm that currently has the highest CTR, we can end the experiment. And one way to achieve that is using Monte Carlo simulation. The way Monte Carlo simulation works is to randomly draw samples from each of the K arms multiple times, and empirically compute how often each of the arms wins (with highest CTR). If the winning arm is beating the second arm by a large enough margin, the experiment terminates. In our experiment, we choose $\alpha = 5\%$, then the experiment terminates when $95\%$ of the samples in a Monte Carlo simulation have remaining value less than $1\%$ of the winning arm's value.

\begin{figure*}[!htp]
\centering
\includegraphics[width=0.9\textwidth]{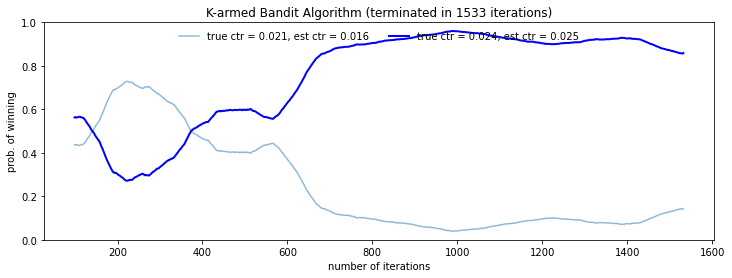}
\caption{2-armed Bandit Test Results}
\label{fig:node5}
\end{figure*}

Figure \ref{fig:node5} shows the 2-armed bandit experiment results (where, arm1 and arm2 are defined earlier in this section). For each iteration, a new user arrives. We apply Thompson sampling to select an arm and see whether user clicks. Then we update the Beta parameters of the arm, check whether we are confident enough about the winning arm to end the experiment. Moreover, we introduced a "burn-in" parameter in this experiment. This is the minimum number of iterations that must be run before declaring a winner. In our experiment we took $burn\_in = 1500$. The beginning of the experiment is the nosiest period, and any loser arm could get ahead by chance. The burn-in period helps prevent prematurely ending the experiment before the noise settles down. As we can see in Figure \ref{fig:node5}, initially the arm 1 was winning, but after some $350$ iterations onwards, arm 2 (dark blue line) becomes the winning arm. The experiment terminates after $1533$ iterations. The amount of traffic sent to arm 1 is $375$ and arm 2 is $1158$. Finally, we conclude from this experiment that the review based nudge/badge seems to be an effective strategy and the e-commerce firm should explore this further by implementing it as a feature and then running an AB test.

In the next experiment, we perform a 3-armed bandit test where the modus-operandi and the parameters remain the same as the previous experiment with the exception of additional variant/arm. The 3 arms are defined as follows:
\\
1) Arm 1- This represents the current experience in the e-commerce firm's website for the concerned category as defined in the previous experiment. The CTR remains $0.021$ as assumed in the previous experiment.   \\
2) Arm 2- This represents the review based nudging strategy as defined in the previous experiment.\\ The CTR is assumed to be $0.024$.   \\
3) Arm 3- This represents a combination of nudges strategy i.e. combination of review based nudge with other trust badges. Here, we take the items with Positive sentiment labels similar to what we did in Arm 2 but having trust badge such as "Bestseller". These trust badges take into account other features such as item transactions, add to carts, and more. We then check the CTR for these set of items from the data and let's it is somewhere close to $0.044$. The CTR here will obviously be a bit higher as these are the premier items in this category. \\

\begin{figure*}[!htp]
\centering
\includegraphics[width=0.9\textwidth]{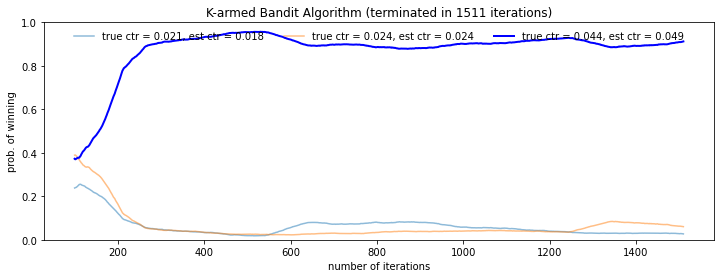}
\caption{3-armed Bandit Test Results}
\label{fig:node6}
\end{figure*}

The goal of the experiment is to find the nudging strategy with the highest click through rate. In Figure \ref{fig:node6}, we show the 3-armed bandit experiment results. We can see that the arm 3 (dark blue line) is clearly the winning arm. The experiment terminates after $1511$ iterations. The amount of traffic sent to arm 1 is $110$, arm 2 is $125$ and arm 3 is $1276$. 

Overall, we conclude from both the above experiments is that the review based nudge/badge seems to be an effective strategy either as a standalone nudge or as a combination with other nudges. It is recommended that the e-commerce firms should explore this further by implementing it as a feature and then running an AB test.

\section{Conclusion} \label{conl}
In this paper, we applied multiple embeddings such as BERT, Glove, and LSTM for Sentiment classification (Positive, Negative, and Neutral) of customer reviews at e-commerce platforms. We then proposed two different Stacking/ensembling approaches. We perform our experiments on three real-world customer reviews data from different e-commerce platforms namely, one internal Walmart dataset and two external \\datasets. We compare the performances of these models using different evaluation metrics such as, prediction accuracy and F1-score. We find that one of the proposed stacked ensemble approach for sentiment classification is the best performer. Finally, we provide some illustrations of how the output from this model can be applied for Nudge marketing by e-commerce firms to help their customers in making purchase decisions. We also performed two k-armed Bandit experiments for the Review based Nudging Strategy and combination of nudges strategy using Thompson sampling and Monte Carlo simulation. From the bandit experiments we conclude that the review based nudge/badge seems to be an effective strategy either as a standalone nudge or as a combination with other nudges. It is recommended that the e-commerce firms should explore this further by implementing it as a feature and then running an AB test.

As a direction for future research, we would like to apply some of the latest techniques for language modeling such as, RoBERTa \citep{roberta} and XLNet \citep{xlnet} on the customer reviews datasets used in this paper to further improve the performance of the Sentiment Classification model.

\bibliography{reviews}

\end{document}